# AI Giving Back to Statistics? Discovery of the Coordinate System of Univariate Distributions by Beta Variational Autoencoder

**Alex Glushkovsky**
BMO Financial Group
alexglu@gmail.com

**Abstract**

Distributions are fundamental statistical elements that play essential theoretical and practical roles. The article discusses experiences of training neural networks to classify univariate empirical distributions and to represent them on the 2D latent space forcing disentanglement based on the inputs of cumulative distribution functions (CDF). The latent space representation has been performed using an unsupervised beta variational autoencoder (β-VAE). It separates distributions of different shapes while overlapping similar ones and empirically realises relationships between distributions that are known theoretically. The synthetic experiment of generated univariate continuous and discrete (Bernoulli) distributions with varying sample sizes and parameters has been performed to support the study. The representation on the latent 2D coordinate system can be seen as an additional metadata of the real-world data that disentangles important distribution characteristics, such as shape of the CDF, classification probabilities of underlying theoretical distributions and their parameters, information entropy, and skewness. Entropy changes, providing an "arrow of time", determine dynamic trajectories along representations of distributions on the latent space. In addition, post β-VAE unsupervised segmentation of the latent space based on weight-of-evidence (WOE) of posterior versus standard isotopic 2D normal densities has been applied detecting the presence of assignable causes that distinguish exceptional CDF inputs.

## 1  Introduction

Statistics plays an essential role in AI by being a fundamental element of all machine learning algorithms. Meanwhile, distributions are key statistical elements that have dual aspects: mathematical equations that support theoretical developments and representation of empirical data. Observing the present "big data" trend, the latest becomes an important aspect for describing data. Unprecedented recent developments in AI and machine learning fields triggers the question: Can we gain more intelligence in statistics itself by applying AI/ML approaches, specifically, concerning univariate distributions?

The immediate application that can be considered is a classification task of recognising the type of underlying distribution observing empirical univariate variables. The traditional and straight forward approach to solve this problem is fitting data against some theoretical distributions and selecting the best fitted one by applying some well-known statistical criteria, such as Kolmogorov-Smirnov (K-S) distance or Kullback–Leibler (K-L) divergence. However, this approach has a strong limitation when dealing with thousands of variables. Machine learning methods can be applied to solve this problem. But, the more intriguing question is "Can we go beyond this trivial task and even learn some insights about distributions?"

In this research, design of experiment (DOE) has been performed by generating variables with known underlying theoretical distributions. Having generated data accompanied by labels, two problems mentioned above have been addressed: first, supervised classification by converting generated variables into CDF and then using simple deep and convolutional neural networks; and second, unsupervised representation of the CDF input on the latent space.

Recently, numerous great approaches have been developed addressing representation on the latent space. It includes embedding, generative variational autoencoders, and disentanglement, such as t-SNE, β-VAE, InfoVAE, InfoGAN (Van der Maaten and Geoffrey, 2008; Bengio, 2013; Kingma and Welling, 2014; Chen *et al*, 2016; Makhzani *et al*, 2016; Higgins *et al*, 2017; Zhao *et al*, 2018).

In this paper, generative β-VAE has been applied considering the regularization term of K-L divergence from standard isotropic multivariate normal distribution that allows for mathematical transformation into a simple form (Kingma and Welling, 2014):

$$\frac{1}{2}\sum_{j=1}^{2}(1 + \log(\sigma_j^2) - \mu_j^2 - \sigma_j^2) \tag{1}$$

where $\mu_j$ and $\sigma_j$ are parameters of the latent space distribution, $j$ – is index of the latent vector (2D in our case).

The unsupervised generative β-VAE has the following features: it compresses information of CDF inputs into a latent 2D vector, disentangles characteristics of the underlying input distributions, and generates output CDF for a given location on the latent space. In addition, the beta parameter, that is a Lagrangian multiplier, provides flexibility to balance between decoded accuracy and disentanglement (Higgins *et al*, 2017).

## 2 Synthetic Experiment

The synthetic experiment has been performed to support the study. Details of the design of experiments (DOE) are shown in Table 1. It includes the generation of one discrete (Bernoulli) and twelve major univariate continuous distributions with varying sample sizes and parameters.

| Id | Distribution Name | Parameters | Type | Domain | Skewness, Cause |
|---|---|---|---|---|---|
| 0 | Beta | α=uniform(0.1, 9.0), β=uniform(0.1, 9.0) | Continuous | (0,1) | Neg/Pos, Parametric |
| 1 | Cauchy | loc=0, scale=1 | Continuous | (-∞,+∞) | Neg/Pos, Stochastic |
| 2 | Exponential | loc=0, scale=1 | Continuous | [0,+∞) | Pos |
| 3 | Gamma | α=uniform(0.1, 9.0) | Continuous | (0,+∞) | Pos |
| 4 | Log-Normal | s=1/uniform(0.1, 9.0) | Continuous | (0,+∞) | Pos |
| 5 | Normal | loc=0, scale=1 | Continuous | (-∞,+∞) | Neg/Pos, Stochastic |
| 6 | Uniform | loc=0, scale=1 | Continuous | [0,1] | N/A |
| 7 | Sup-Normal | loc1=0, scale1=1, loc2=uniform(0, 1), scale2=uniform(0.1, 9.0), w=uniform(0.1, 0.9) | Continuous | (-∞,+∞) | Neg/Pos, Parametric |
| 8 | Weibull | α=uniform(0.1, 9.0) | Continuous | [0,+∞) | Neg/Pos, Parametric |
| 9 | Chi | df=random(1,2,3,… 9) | Continuous | [0,+∞) | Pos |
| 10 | Bernoulli | p=uniform(0.001, 0.999) | Discrete | {0,1} | Neg/Pos, Parametric |
| 11 | Gumbel_L | loc=0, scale=1 | Continuous | (-∞,+∞) | Neg |
| 12 | Gumbel_R | loc=0, scale=1 | Continuous | (-∞,+∞) | Pos |

Table 1. Simulated DOE

Overall, 117,000 variables (9,000 per each distribution type) have been generated using the "scipy.stats" package in Python. Sample sizes of the generated random series are varying randomly from 35 to 1,000 supporting robustness of conclusions in practical environments.

Selection of distribution types have been limited to the basic well-known and widely used distributions fitting real-world data (Leemis and McQueston, 2008). In addition, synthetic distribution "Sup-Normal" (#7, Table 1) has been constructed superimposing two normal distributions.

Generated random variables have been converted into cumulative distribution functions (CDF) based on two grids (16x15) and (26x25), where the first number is the number of grouping bins along the X-axis and the second number is the number of levels on the Y-axis of the cumulative function. To increase sensitivity of the representation having a discrete number of grouping bins, the modified CDF includes a third dimension: the number of observations that fall into a specific cell of the grid that has been scaled to [0, 1] range.

It turns out that both (16x15) and (26x25) grids provide quite comparable results, although the more granular grid (26x25) produces more smooth modeling results while not being notably affected by small sample sizes. Therefore, all following discussions will be considering only that grid.

Examples of the modified CDF representations are shown in Figure 1.

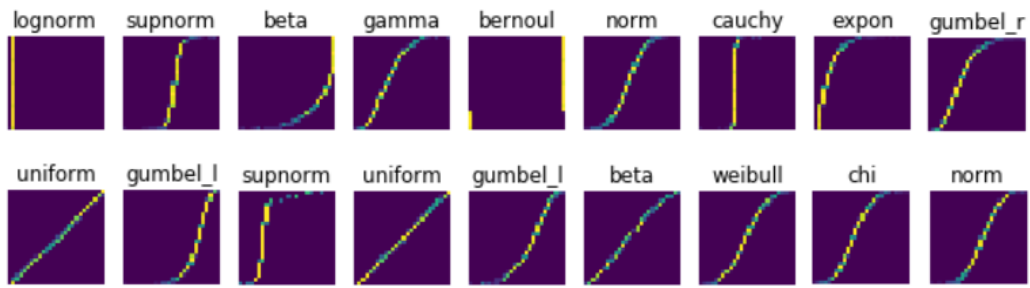

Figure 1. Examples of the generated CDFs

Parameters that change shape of the distribution have been randomly varying within wide ranges (see Table 1). For distributions with fixed parameters loc=0 and scale=1, the random generation still produces fluctuated CDFs due to the nature of the stationary but yet a stochastic process and occurrence of outliers.

To construct CDF, the rank function has been used for values of generated series. There are three different methods returning the rank number in Python: "average", "first", and "max". To provide consistency with continuous distribution cases, the "first" method has been used in the experiment for Bernoulli distribution.

Modified CDF provides a universal representation of distributions having diverse types, a wide variety of parameters, and different ranges of generated values. It produces arrays that can be seen as standardized images of 26x25 "pixels" (see Figure 1). The latest allows for usage of the well-developed image recognition models, such as convolutional neural networks.

## 3 Characteristics of the Generated Distributions
### 3.1 Information Entropy

Normalized information entropy for each generated random variable has been calculated based on Shannon's discrete formula (Shannon, 1948):

$$E = -\sum_{i=0}^{N_{BINS}} P_i * log_2(P_i) / log_2(N_{BINS}) \quad (2)$$

where $N_{BINS}$ is number of bins and in our case, it equals 26.

It allows for comparison of information entropies between different types of distributions having a universal scale between zero and one.

### 3.2 Non-Parametric Skewness Measure

To study skewness of the generated distributions, a non-parametric approach has been applied based on signed Kolmogorov-Smirnov distances to the theoretical Uniform distribution:

$$Skw = D_{pos} - D_{neg} \quad (3)$$

where $D_{pos,neg}$ – are maximum Kolmogorov-Smirnov distances when the CDF is above or below the diagonal line, correspondingly (Filion, 2015).

### 3.3 Kolmogorov-Smirnov Distance to Uniform Distribution

To analyze differences between CDF shapes of two distributions, the maximum Kolmogorov-Smirnov distance has been applied. Instead of performing a direct pairwise comparison between two distributions, the Uniform CDF has been used as a reference (i.e., the diagonal line). It allows for significant reduction of the computational complexity from $O(N^2)$ to $O(N)$.

## 4 Distribution Type Recognition by Neural Networks

To classify distributions by type, two simple supervised models have been trained: (1) deep sequential, and (2) convolutional neural network. Architecture of these models are similar to basic MNIST classification models (https://keras.io/examples/mnist_mlp/ and https://keras.io/examples/mnist_cnn/).

Both models provide quite similar results at around 79% accuracy for the testing dataset, and, therefore, the results of the first simpler model will be discussed further.

| | | Predicted | | | | | | | | | | | | | |
|---|---|---|---|---|---|---|---|---|---|---|---|---|---|---|---|
| | | Beta | Cauchy | Expon | Gamma | Log-Norm | Normal | Uniform | Weibull | Sup-Norm | Chi | Bernoulli | Gumbel-R | Gumbel-L | Total | Accuracy |
| Generated True | Beta | 1781 | 1 | 46 | 192 | 18 | 159 | 22 | 211 | 54 | 369 | | 6 | 90 | 2949 | 60.4% |
| | Cauchy | | 2905 | | | 12 | | | | 20 | | 11 | 3 | 5 | 2956 | 98.3% |
| | Expon | 17 | | 2868 | 92 | 20 | | 1 | | 5 | | | | | 3003 | 95.5% |
| | Gamma | 66 | | 139 | 1861 | 155 | 3 | | 14 | 10 | 147 | | 589 | | 2984 | 62.4% |
| | Log-Norm | 11 | 35 | 23 | 497 | 1251 | 52 | 2 | 25 | 26 | 582 | | 445 | 1 | 2950 | 42.4% |
| | Normal | 120 | | | 11 | 40 | 2371 | | 196 | 55 | 213 | | 6 | 12 | 3024 | 78.4% |
| | Uniform | 7 | | | 2 | | 1 | 2913 | 1 | 4 | 1 | | | 1 | 2930 | 99.4% |
| | Weibull | 301 | 8 | 72 | 377 | 87 | 278 | 5 | 1371 | 33 | 328 | | 11 | 104 | 2975 | 46.1% |
| | Sup-Norm | 31 | 8 | 1 | 11 | 33 | 73 | 4 | 17 | 2651 | 46 | | 19 | 15 | 2909 | 91.1% |
| | Chi | 121 | | 7 | 275 | 313 | 150 | 3 | 26 | 22 | 1995 | | 36 | 2 | 2950 | 67.6% |
| | Bernoulli | | 8 | | | | | | | | | 2985 | | | 2993 | 99.7% |
| | Gumbel-R | 3 | | 2 | 378 | 60 | 6 | 2 | | 10 | 21 | | 2485 | | 2967 | 83.8% |
| | Gumbel-L | 32 | | | | 16 | 4 | 57 | 6 | 1 | | | | 2904 | 3020 | 96.2% |

Table 2. Confusion matrix

The accuracy of classification is affected by theoretical overlapping, for example between Gamma and Exponential distributions, and empirical cases having almost identical CDFs that belong to different types of generated distributions.

To learn features that impact quality of classification, a shallow decision tree has been trained. It turns out that most confusions occur between Beta, Chi, Gamma, Log-Normal, and Weibull types of distributions (Table 2). Also, having sample sizes of less than 220 observations to form CDF increases misclassification for these types of distributions.

The confusion matrix indicates that the trained neural network tends to predict simpler (less parametrizes) distributions – only in 6% of cases it was confused with more parametrized types of distribution.

Using a trained model against real-world data classifies the type of underlying distribution and can be added in the descriptive summary of the variable along with count of observations, mean, variance, etc., i.e., to be part of the metadata of the collected data. In addition, it can be accompanied by parameter estimations that are suitable to the specific type of classified distribution.

Another important outcome of the trained model is distinction between underlying distributions that is provided by the SoftMax activation function of the final layer. The output values of the SoftMax function can complement the metadata describing input variables.

Furthermore, the described classification model can be helpful in identifying outliers in univariate cases. Knowing the type of underlying distribution is an important input for detecting outliers.

## 5  Application of Beta Variational Autoencoder (β-VAE) for CDF Inputs

Unsupervised application of β-VAE for generated CDF has been explored to (1) encode information of 650 (26x25) inputs into a very limited number of latent variables of the "bottleneck" layer (2D in our case) and then to decode them by restoring the initial input variables; (2) disentangle CDF representation on the latent space by tuning the beta multiplier; and (3) to generate CDF for a specified location on the latent space (https://keras.io/examples/variational_autoencoder/).

In addition, a supervised deep neural network has been trained to classify the distribution type by having latent vector as inputs.

### 5.1  Latent Space as a Coordinate System of Univariate Distributions

Distribution mapping on the latent space by applying β-VAE represents a coordinate system of the univariate distributions (Figure 2).

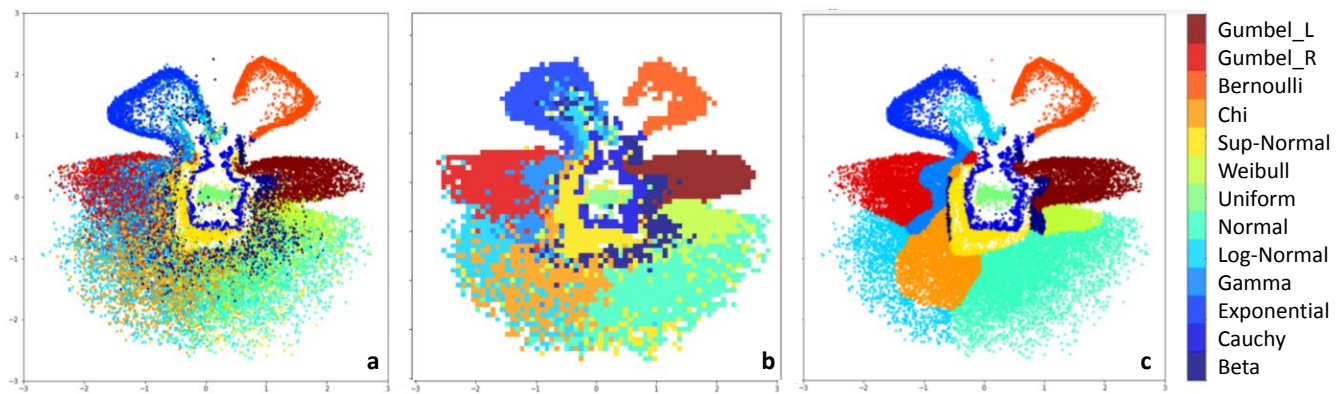

Figure 2. Example of posterior distribution on the latent space (β=3): (a) original CDF test data; (b) applying maximum distribution type representation on 75x75 grid; (c) fitting classification neural network model providing latent vector as input

The empirical unsupervised representation in Figure 2 is well aligned with theoretical knowledge about underlying distributions. Thus, Uniform distribution has a maximum information entropy for standardized ranges across all distributions. Unsupervised β-VAE neural network has placed this distribution around the central point on the latent space. This point can be seen as a "black hole" of information. It is surrounded by Cauchy distribution forming a crater of quite low information entropy values that are caused by heavy tails and outliers. The most concentrated distributions on the latent space are: Uniform, Bernoulli, Cauchy, and Exponential. Notably Gumbel_L and Gumbel_R distributions are placed quite symmetrically on the latent space.

## 5.2 Characteristics of Distributions on the Latent Space

Some characteristics of the underlying distributions are presented in Figure 3. The information entropy surface has a crater-like topology on the bottom half and gradients along the outlines of the top half of the latent space. Distributions are well disentangled by skewness, where red points represent positive and blue points represent negative skewness, correspondingly. Prevalence of positive skewness is caused by un-balanced DOE. The highest divergence from Uniform, which is measured by $D_{K-S}$ distance, can be observed on the "crater" (Cauchy) and the top-left outline (Exponential, Log-Normal).

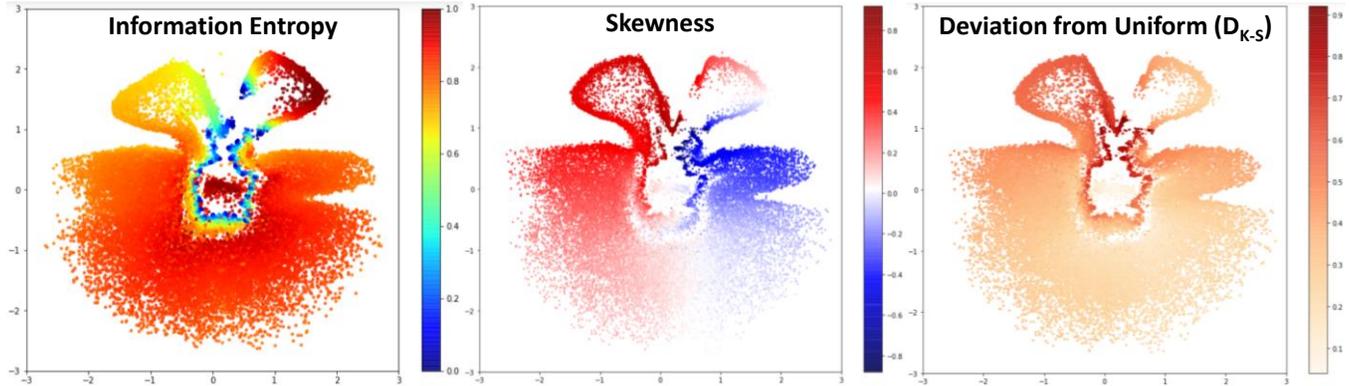

Figure 3. Mapping of CDF characteristics on the latent space

In addition, parameters of underlying distributions can be mapped on the latent space. Example for $\alpha$ parameter of Weibull distribution is shown in Figure 4, a. Clearly, there is a distinct pattern of $\alpha$ parameter representation.

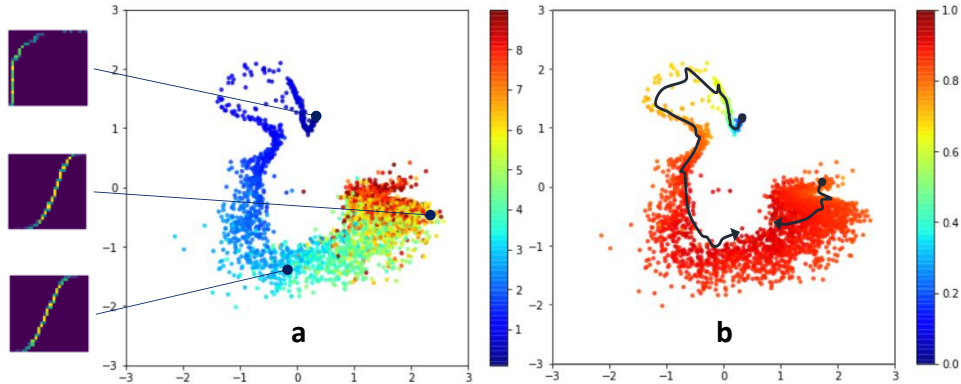

Figure 4. Example of Weibull distribution characteristics on the latent space: (a) distribution parameter α, (b) information entropy with trajectories along "arrow of time",

## 5.3 Distribution Representation on the Latent Space as a "Dynamic" System

Acknowledging that an entropy increase represents an "arrow of time" (Eddington, 1928), it can be applied to deriving average trajectories from lowest to highest information entropy values for distributions on the latent space (see Figure 5).

Number of trajectories per distribution depends on skewness variation of the generated CDF. For example, Exponential distribution has a single trajectory, whereas Weibull has dual starting points that are caused by $\alpha$ parameter changes, and Normal has dual trajectories that are caused by stochastic deviations.

It turns out that (1) Exponential, Gamma, Weibull, and Log-Normal distributions have very close origination points, (2) the end of the Exponential distribution is very close to the origination point of Chi, and (3) the ends of Beta and Uniform distributions are close to the central point while the other trajectories (except Bernoulli) are ending up towards the "crater".

In addition, theoretical equalities between Exponential, Gamma, and Weibull distributions at certain parameters, are retained and can be seen as intersections in the shaded zone A. Similarly, Beta, having $\alpha$ and $\beta$ parameters equal one, becomes Uniform (zone B).

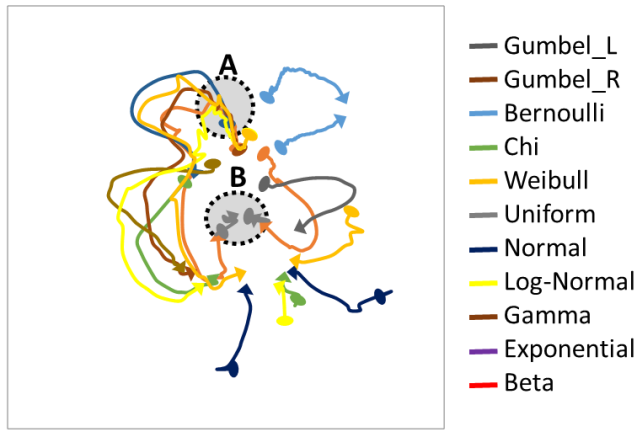

Figure 5. Trajectories along entropy increase ("arrow of time") on the latent space

The spread around trajectories increases significantly as information entropy increases but then shrinks down when information entropy approaches the maximum value (Figure 4, b). It is caused by the generation of the same number of variables per distribution types in the DOE that appear with a higher concentration of their entropies within (0.7:0.9) range. This issue can be addressed in future research by re-balancing distributions by their characteristics.

### 5.4 Relationships between Distributions on the Latent Space

Relationships between distributions on the latent space are shown in Figure 6.

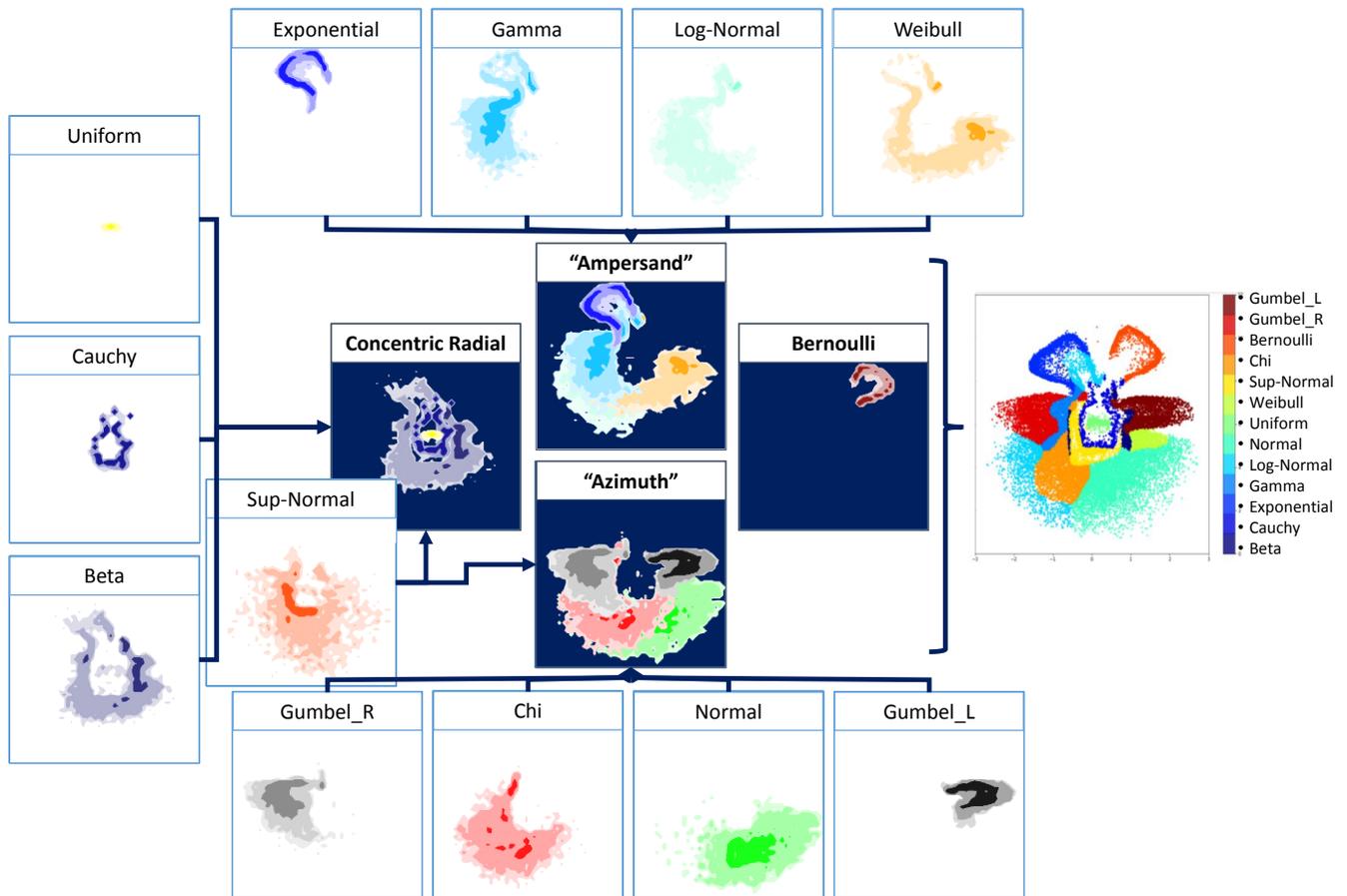

Figure 6. Relationships between distributions on the latent space.

The unsupervised representation on the latent coordinate system forms four distinct associations: (1) concentric radial (Uniform, Cauchy, and Beta); (2) "azimuth" (Gumbel_R, Chi, Normal, and Gumbel_L); (3) "ampersand" (Exponential, Gamma, Weibull, and Log-Normal); and (4) stand along Bernoulli. It can be noted that considering only the first two associations, the orthogonal 2D latent space can be transformed into a polar coordinate system.

The diagram shown in Figure 6 reveals intersections within each association group as well as between them. These observations make sense considering underlying theoretical distributions (Leemis and McQueston, 2008). Thus, Exponential distribution is extended by Gamma following by Weibull and Log-Normal. Gumbel_L distribution is bordered with Normal. The latest is extended by Chi with some intersection, followed by Gumbel_R. Being a discrete distribution, Bernoulli is almost perfectly isolated except for one point ($p\rightarrow 1$) touching Beta with $\alpha \gg 1$, $\beta \rightarrow 0$ and Cauchy that extremely skewed to the left. In addition, synthetic Sub-Normal distribution lays between concentric radial and "azimuth" groups. It is supported by the fact that depending on the parameters, this distribution may match either Normal, dual modal, or even Uniform.

## 6  CDF Generation by β-VAE

Even though the DOE covers a wide spectrum of distributions (see Table 1 and Figure 1), there are empty areas on the latent space. Generative VAE can produce outputs for any location including empty spaces on the latent space (Kingma and Welling, 2014). Examples of such generation are shown in Figure 7, a. It can be seen as a synthesis including new types of distributions. The β-VAE generates new CDF by interpolating "images" from existing surrounding points. However, in this approach, the principal property of any CDF to be a monotonic increase function can be lost.

That property can be reinstated by running a constraint optimization task after β-VAE generation ensuring monotonity of the restored CDF, while minimizing deviations between generated and corrected functions. Example of such generation with the following constrained fitting is shown in Figure 7, b.

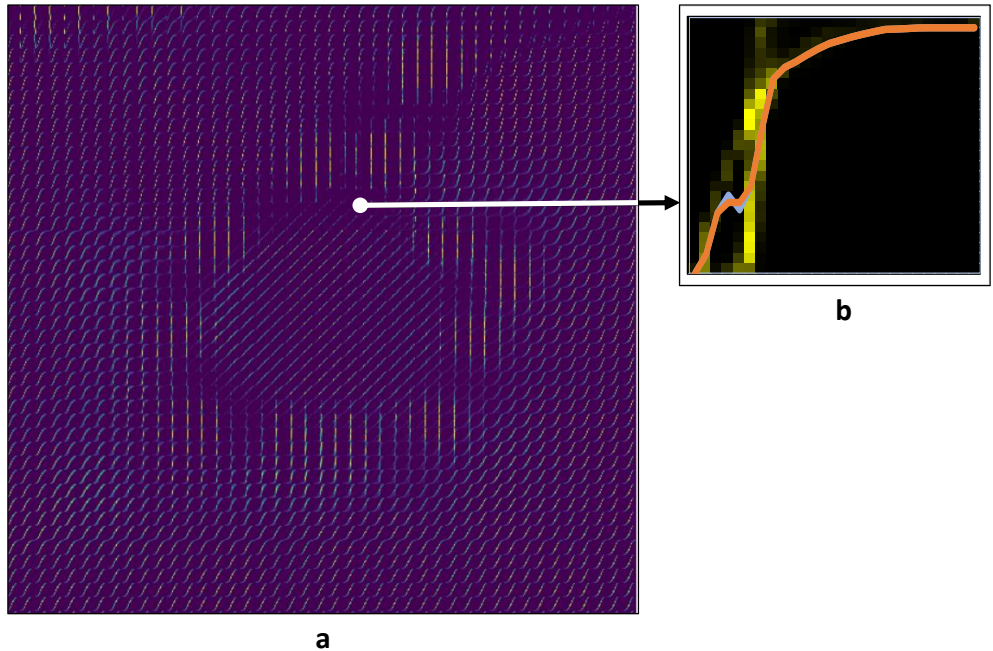

Figure 7. (a) Grid of generated CDFs by β-VAE on the latent space (50x50); (b) example of the generated CDF for the empty space, where blue line: the weighted function; orange line: CDF after constrained fitting

## 7  Post β-VAE Unsupervised Segmentation on the Latent Space

The loss function of β-VAE consists of two parts. The first part is responsible for autoencoder accuracy to restore inputs and the second part forces the posterior distribution to be isotropic normal by applying the K-L regularization term (Kingma and Welling, 2014; Higgins *et al*, 2017).

Considering a bivariate posterior probability density of the latent variables $p(L_1, L_2)$, it can be observed that it deviates from the standard 2D normal shape (Figure 8). It means that the first part of the loss function is dominant in some regions and that there are assignable causes affecting inherent structure of corresponding inputs there.

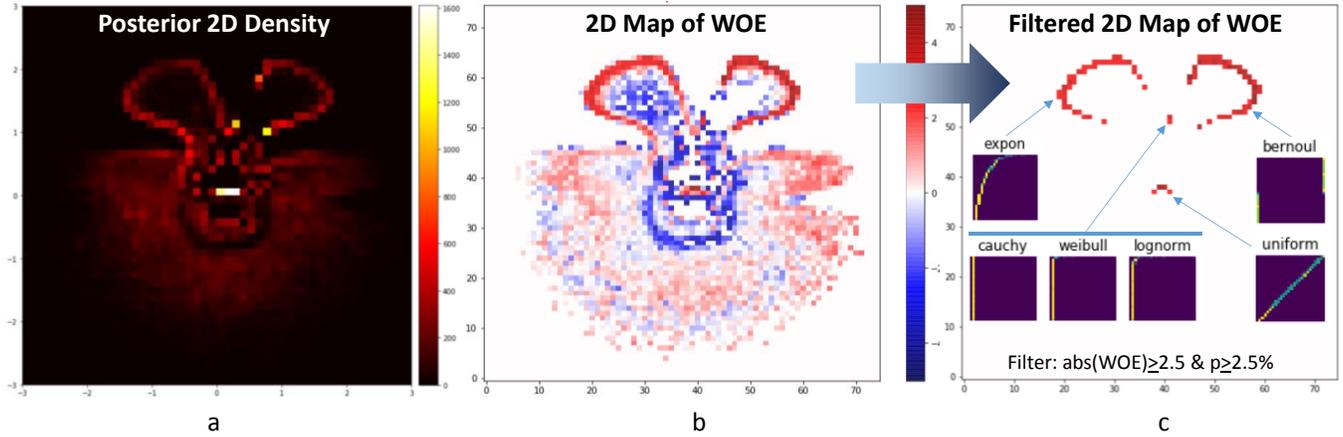

Figure 8. Posterior density and map of WOE on the latent space

To explore this issue, the weight of evidence (WOE) ratio has been applied (Good, 1986):

$WOE(L_1, L_2) = \log(p(L_1, L_2))/\log(2D\_N(0,1)) = \log(p(L_1, L_2)) - \log(2\pi) - 0.5(L_1^2 + L_2^2)$ (4)

where $L_{1,2}$ are coordinates of the latent space.

Based on the calculated WOE, the latent space can be split into segments where the fitted posterior distribution density is significantly different than standard isotropic 2D normal. To illustrate this approach, the following splitting criteria has been applied $abs(WOE) > W^*$ and $p(L_1, L2) \geq 0.025$, where $W^* = 2.5$ is an arbitrary threshold value (Figure 8, c). The splitting criteria may vary and, for example, a univariate decision tree can be used for segmentation.

Unsupervised post β-VAE segmentation, which is illustrated in Figure 8, c, differentiates four isolated areas from more homogenous common space. It points to the existence of assignable causes that discriminate input CDFs there. These four areas correspond to Uniform, Bernoulli, Exponential, and the extremely right skewed Cauchy, Log-Normal ($s \rightarrow 0$), and Weibull ($\alpha \rightarrow 0$) distributions. These distributions are specific with respect to their shapes and underlying characteristics compared to the rest of the distributions, but it is also a reflection of the unbalanced DOE towards positively skewed distributions (Table 1).

## 8   Views on Practical Applications

Practical implementations of the discussed univariate empirical distributions classification and their representation on the 2D latent space may include:

- Metadata of the standardized 2D latent space encoding, its graphical representation and characteristics, such as prediction of underlying types of distributions, corresponding parameters estimation, information entropy, K-S distance to Uniform distribution, and non-parametric skewness. This is especially helpful when dealing with "big data" and sophisticated databases. Encoded metadata enhances basic statistics and variable descriptions. It could potentially be used for hyper learning and control on a database metadata level.
- Outlier detection attributed to a specific CDF, for example, concerning Cauchy distribution.
- Simulations and stress tests that include variance of CDF. Usually, stress tests assume some shifts in parameters, such as mean or expected shortfall, but fixed type of underlying distribution. Considering changes of the CDF shapes may provide important insights playing more holistic scenarios. That can be achieved by locating the position with required characteristics on the latent space with the following CDF decoding by β-VAE and simulation of Min and Max values. Knowing the normalized CDF and [Min, Max] range allows for random generation of the underlying variable.
- Periodic monitoring of movements on the latent space allows for control of the CDF shape and corresponding characteristics. It can complement the traditional parametric statistical process control (Montgomery, 2013).

## 9   Real-World Example

The trained model has been applied to five air pollutant emissions (sulphur oxides, nitrogen oxides, carbon monoxide, volatile organic compounds and fine particulate matter) of 36 countries measured in 2007 and 2017 with some missing records (source: Canadian Environmental Sustainability Indicators). These measures are presented as total values and as normalized ratios by GDP.

Encoding of the input CDFs on the latent space of these air pollutant emissions are presented in Figure 9. It can be observed that there is a separation of the points representing total values and normalized ratios. The latest have significantly higher entropies (on average 0.61 versus 0.41) meaning that normalized measures are less affected by outliers and are classified as Log-Normal distribution, while total values are mostly recognized as Cauchy. This aligns well with the rationale for normalization.

Changes over the decade, between 2007 and 2017, are presented as arrows. Interestingly, they are mostly shifted toward higher entropies for normalized measures while having almost no change to total values.

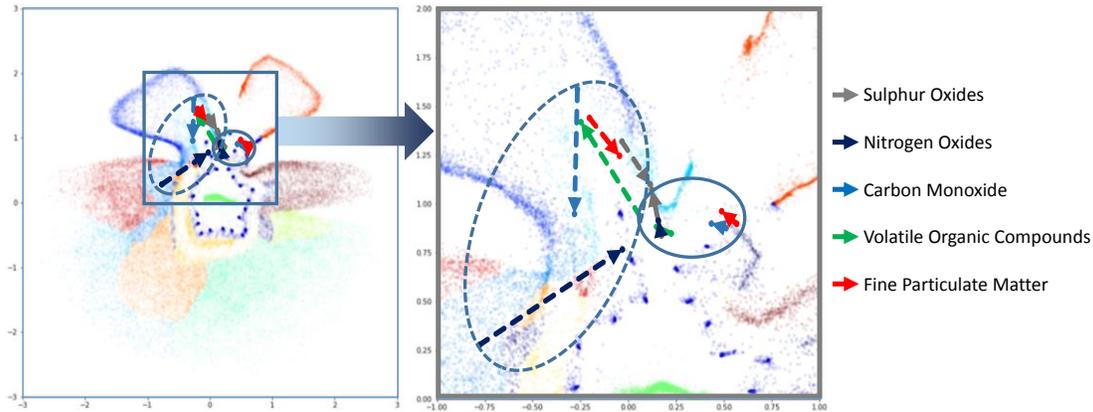

Figure 9. Representation of CDF of air pollutant emissions on the latent space. Solid lines: total measures; dashed lines: normalized ratios

## 10　Extreme Representation on One-Dimensional Space

Examples of representations and decoding on one-dimensional latent space by β-VAE (β=3) are shown in Figure 10.

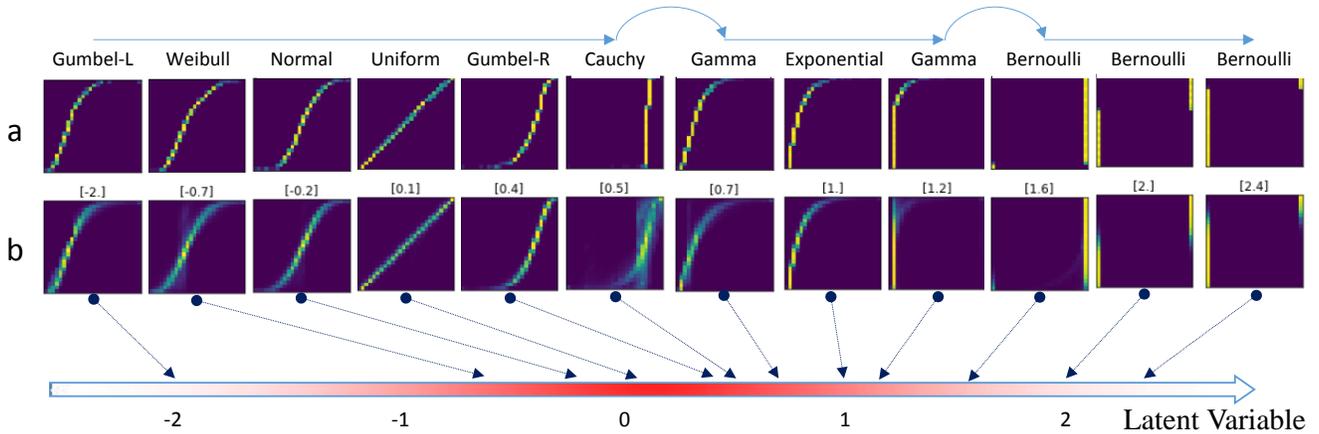

Figure 10. Examples of 1D variational autoencoding along the latent variable: (a) input CDFs; (b) decoded CDFs

As expected, both decoding and disentangling qualities have been deteriorated compared to the 2D case. The latest can be observed in Figure 11, a, which presents posterior densities by distribution types along the latent variable. For some types of distributions, their representations are not continuous along the latent variable (Cauchy, Gamma, Log-Normal, and Weibull). The separation occurs around the central point where Uniform distribution has been placed by the unsupervised neural network. In addition, information entropy along the latent variable has no clear pattern. It has been impacted by poor disentanglement of Cauchy distribution.

Nevertheless, even one-dimensional representation reveals some meaningful relationships between distributions shown in Figure 11, b. For example, Bernoulli is isolated again touching only Beta and Cauchy. Disentanglement of CDF shapes has smooth transitions except for two bifurcations. First, when the latent variable value is around 0.4 by jumping from the extremely left skewed shapes (Cauchy) to the right ones, and, second, around the latent variable value of 1.4 by jumping from the extremely right skewed distributions (Gamma, Log-Normal) to Bernoulli with $p \rightarrow 1$. Both bifurcation points can be detected by described above unsupervised WOE approach as having significant deviations from $N(0,1)$ curve (Figure 11, a).

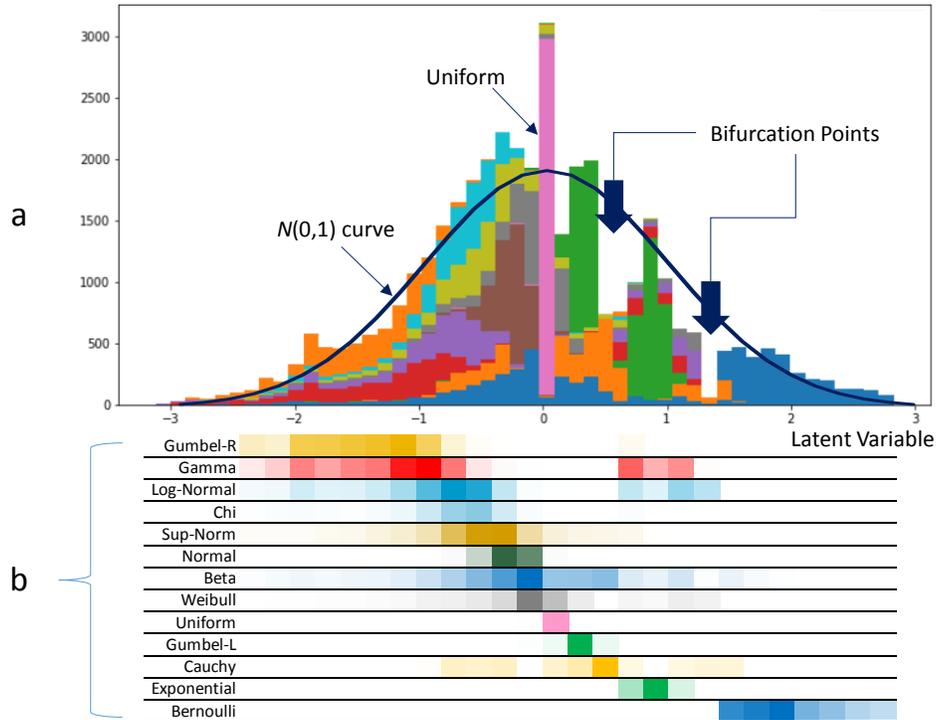

Figure 11. Example of one-dimensional space representation: (a) posterior densities by distribution types; (b) associations between distributions

## 11   Future Research

Future research may include DOE extension and re-balancing (Table 1) incorporating more types of distributions and wider ranges of parameters, as well as applying different disentangled approaches, model architectures, and tuning. For example, representation of types of continuous and discrete distributions is a good candidate to be studied by applying the JointVAE model (Dupond, 2018).

Applied DOE includes generation of variables with known underlying distributions followed by convergence into CDF. Alternatively, CDFs that are inputs of β-VAE can be directly generated applying only a monotonic increase constraint while ensuring more incremental and balanced CDF shapes between extremely left and right skewed cases. This synthetic unlabelled CDF generation will lead to more consistent representation on the latent space having less irregularities such as empty areas.

Trajectories on the latent space that are presented in Figure 5 can be modeled as a system of ordinary dynamic differential equations:

$dL_{1,2}/dE = f_{1,2}(L_1, L_2)$

It can be seen as an analogy to Lorenz strange attractors (Lorenz, 1963), however, this topic requires further research.

## 12   Conclusions

Application of neural networks classification and the beta variational autoencoder allow for prediction of underlying distribution type and unsupervised representation of distributions. The representation comprises mapping of CDF on the universal latent 2D coordinate system. It separates distributions of different shapes while overlapping similar ones and empirically realises relationships between distributions that are known theoretically.

The 2D encoding can be seen as additional metadata of the real-world data that disentangles important distribution characteristics, such as shape of the CDF, information entropy, probabilities of underlying theoretical distributions, and their parameters and skewness.

Post β-VAE unsupervised segmentation of the latent space based on weight-of-evidence (WOE) of posterior versus standard isotopic 2D normal densities can be applied detecting the presence of assignable causes that distinguish exceptional CDF inputs.

Entropy changes, providing an "arrow of time", determine dynamic trajectories along representations of distributions on the latent space.

## Disclaimer

The paper represents the views of the author and do not necessarily reflect the views of the BMO Financial Group.

## Appendix

Generated random variables have been converted into cumulative distribution functions (CDF) by taking the following steps:

1. Scaling of the original variable values to [0,1] range
2. Grouping of the scaled values into 26 bins along the X-axis of the CDF
3. Ranking of the scaled values into 25 levels along the Y-axis of the CDF using ranking method "first"
4. Counting the number of observations that fall into a specific cell of the (26x25) grid
5. Scaling the count of p.4 to [0, 1] range representing normalized Z-axis values of the CDF

Architectures of the neural network models used in this research have been as follows.

Classification models:

> <u>Deep sequential</u>: Flatten input with 650 (26x25) channels; Dense (128, activation='relu'); Dense (64, activation='relu'); Dense (num_classes=13, activation='softmax'); loss= categorical_crossentropy; optimizer=rmsprop; metric=accuracy.
>
> <u>Convolutional neural network</u>: Input 650 (26x25); Conv2D (32, kernel_size=(3, 3), activation='relu'); Conv2D (64, kernel_size=(3, 3), activation='relu'); MaxPool2D (pool_size=(3, 3)); Dropout (0.25); Flatten; Dense (128, activation='relu'); Dense (num_classes=13, activation='softmax'); loss= categorical_crossentropy; optimizer=Adadelta; metric=accuracy.

β-VAE model:

> <u>Encoder</u>: Flatten input with 650 (26x25) channels; Dense (512, activation='relu'); Dense (64, activation='relu'); two Dense (latent_dim=2).
>
> <u>Decoder</u>: Dense (64, input_dim=2, activation='relu'), Dense (512, activation='relu'), Dense (original_dim=650 (26x25), activation='sigmoid').
>
> Loss=binary_crossentropy+beta*KL; optimizer=rmsprop.

Classification model based on the latent vector input:

> <u>Deep sequential</u>: Flatten input with 2 channels; Dense (1024, activation='relu'); Dense (64, activation='relu'); Dense (num_classes=13, activation='softmax'); loss= categorical_crossentropy; optimizer= Adadelta; metric=accuracy.

Split between training and testing datasets has been 67%/33% for all models.